%% file: camera_ready.tex
\documentclass[10pt,twocolumn,letterpaper]{article}

\usepackage{iccv}              % To produce the CAMERA-READY version
\usepackage[accsupp]{axessibility}

\definecolor{iccvblue}{rgb}{0.21,0.49,0.74}
\usepackage[pagebackref,breaklinks,colorlinks,allcolors=iccvblue]{hyperref}

\def\paperID{11361	} % *** Enter the Paper ID here
\def\confName{ICCV}
\def\confYear{2025}

\title{Adaptive Articulated Object Manipulation On The Fly with Foundation Model Reasoning and Part Grounding}

\author{Xiaojie Zhang\textsuperscript{*\rm 1}, Yuanfei Wang\textsuperscript{*\rm 2}, Ruihai Wu\textsuperscript{*\rm 2}, \\Kunqi Xu\textsuperscript{\rm 3}, Yu Li\textsuperscript{\rm 1}, Liuyu Xiang\textsuperscript{\rm 1},
Hao Dong\textsuperscript{\dag \rm 2}, Zhaofeng He\textsuperscript{\dag \rm 1}\\
\textsuperscript{1} Beijing University of Posts and Telecommunications \\
\textsuperscript{2} School of Computer Science, Peking University 
\textsuperscript{3} School of EECS, Peking University
}

\begin{document}
\maketitle
\input{sec/0_abstract}    
\let\thefootnote\relax\footnotetext{*Equal contribution. \dag Corresponding authors. }
\input{sec/1_intro}

\input{sec/2_related}
\input{sec/3_method}

\input{sec/4_experiments}

\input{sec/5_conclusion}
\input{sec/6_ack}
{
    \small
    \bibliographystyle{ieeenat_fullname}
    \bibliography{main}
}

\end{document}

%% file: sec/0_abstract.tex
\begin{abstract}
Articulated objects pose diverse manipulation challenges for robots. 
Since their internal structures are not directly observable, robots must adaptively explore and refine actions to generate successful manipulation trajectories. While existing works have attempted cross-category generalization in adaptive articulated object manipulation, two major challenges persist: (1) the geometric diversity of real-world articulated objects complicates visual perception and understanding, and (2) variations in object functions and mechanisms hinder the development of a unified adaptive manipulation strategy.
To address these challenges, we propose \textbf{AdaRPG}, a novel framework that leverages foundation models to extract object parts, which exhibit greater local geometric similarity than entire objects, thereby enhancing visual affordance generalization for functional primitive skills. To support this, we construct a part-level affordance annotation dataset to train the affordance model. Additionally, AdaRPG utilizes the common knowledge embedded in foundation models to reason about complex mechanisms and generate high-level control codes that invoke primitive skill functions based on part affordance inference.
Simulation and real-world experiments demonstrate AdaRPG’s strong generalization ability across novel articulated object categories.

\end{abstract}

%% file: sec/1_intro.tex
\vspace{-0.3cm}
\section{Introduction}
\vspace{-0.1cm}
\label{intro}
\input{fig-tab/fig1}
Articulated object manipulation is a fundamental challenge in robotics, requiring robots to interact with objects composed of multiple interconnected parts and joints. Real-world tasks often demand adaptive manipulation, where robots must handle realistic, long-horizon interactions by reasoning about an object's functional mechanisms. Adaptive manipulation requires robots to infer hidden states and adjust their actions accordingly. For example, a safe can only be opened if its latch is unlocked—yet the lock state is invisible to the robot, forcing it to attempt pulling the handle, observe feedback, and adapt its strategy. This adaptability is essential for autonomous robots operating in dynamic, unstructured environments, where object behaviors are not explicitly predefined.

Despite extensive research, cross-category generalization in adaptive articulated object manipulation remains a major challenge due to two key factors. First, geometric diversity among real-world articulated objects complicates visual perception and affordance learning. Traditional visual affordance models struggle to generalize across novel object categories, limiting their ability to transfer functional skills. Second, variation in manipulation mechanisms—including joint constraints, locking mechanisms, and randomized rotations—prevents direct policy transfer between object categories. These factors create a significant barrier to developing a unified adaptive manipulation policy that works across diverse articulated objects.

Fortunately, recent advances in foundation vision and language models have demonstrated strong generalization capabilities in grounding and reasoning. Trained on vast datasets, these models present a promising opportunity to enhance robotic perception and mechanism reasoning. 
At the same time, articulated objects consist of multiple functional parts, with manipulation trajectories inherently determined by these parts. Moreover, different object categories often share similar parts, making part-based representations a natural and effective choice for articulated object manipulation.
Building on these insights, we introduce \textbf{Ada}ptive Manipulation with Foundation Model \textbf{R}easoning and \textbf{P}art \textbf{G}rounding (\textbf{AdaRPG}), a novel framework that utilizes foundation models to achieve generalizable adaptive articulated object manipulation. 

As shown in \cref{fig:fig1}, we first construct a part affordance dataset, consisting of part-level point clouds and corresponding affordance annotations. During inference, AdaRPG grounds and segments object parts to extract the relevant part point cloud, which is then used to infer part affordance for guiding primitive skill functions. Since object parts share greater local geometric similarity than whole objects, this approach significantly improves affordance generalization to novel objects. Next, AdaRPG leverages pre-trained knowledge from GPT-4o to reason about manipulation mechanisms and generate control codes that invoke primitive skill functions for execution. By integrating foundation model reasoning, AdaRPG can automatically generalize to unseen object categories and effectively execute long-horizon adaptive policies.

To validate our proposed framework, we conduct experiments on seven categories of simulated objects and four categories of real-world objects, each exhibiting adaptive mechanisms. Through both simulation and real-world evaluations, we demonstrate that AdaRPG significantly improves cross-category generalization, outperforming existing methods in handling novel articulated objects.

In summary, our contributions include:

\begin{itemize}
    \item We introduce a foundation model-powered framework that enhances generalization via part segmentation and mechanism reasoning for control code generation.
    \item We annotate a part-level affordance dataset to support generalizable affordance model guiding primitive skills.
    \item Extensive simulation and real-world experiments demonstrate the effectiveness of our AdaRPG framework.
\end{itemize}

%% file: fig-tab/fig1.tex
\begin{figure*}
  \centering
  \includegraphics[width=0.95\textwidth]{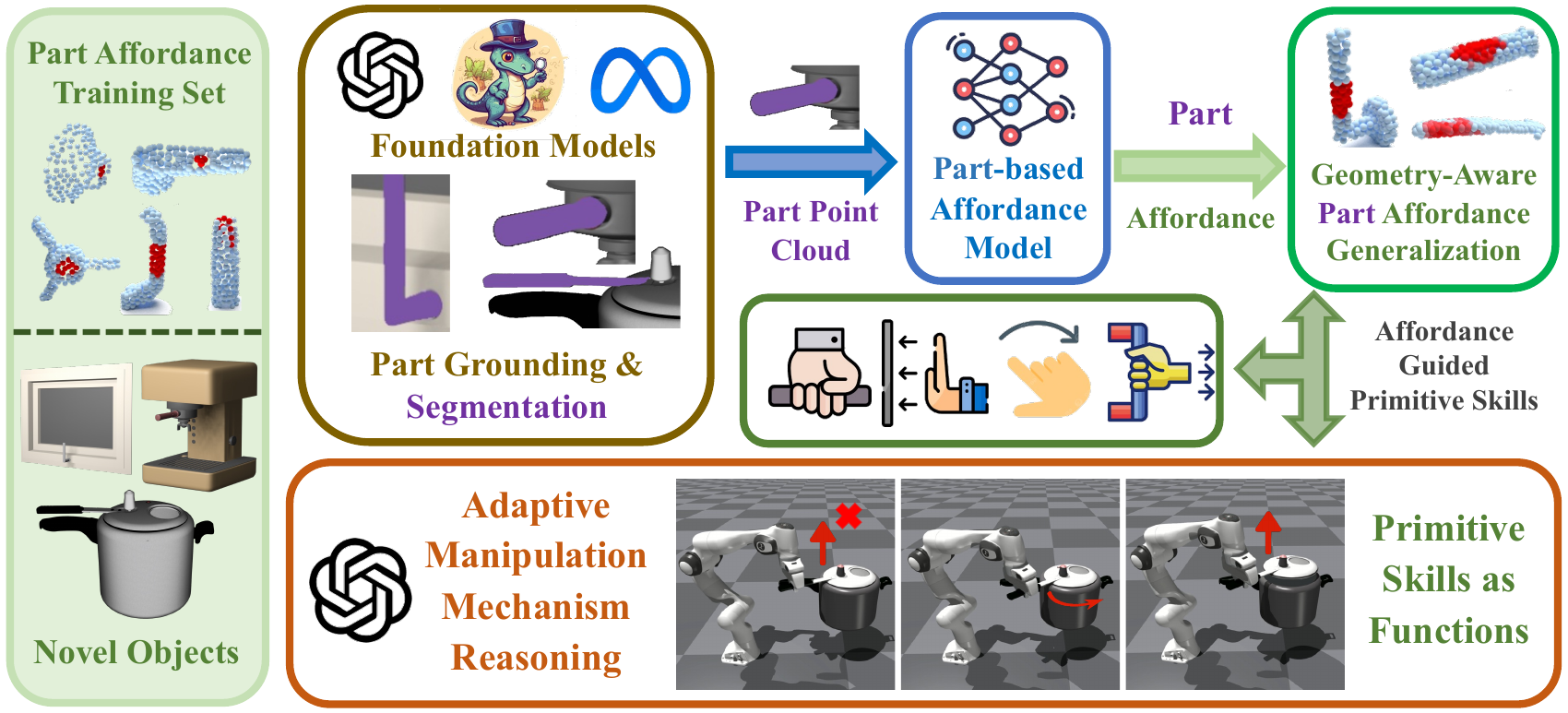}
  \caption{Overview of \textbf{AdaRPG}. We construct a part-level affordance dataset to train the affordance model, which takes a part point cloud as input. During inference, AdaRPG first leverages foundation vision and language models to segment the part point cloud and predict part affordance, guiding primitive skills such as grasping and rotating. AdaRPG then utilizes GPT-4o to reason about manipulation mechanisms and generate Python code that invokes primitive skill functions for execution.
  }
  \label{fig:fig1}
  \vspace{-0.3cm}
\end{figure*}

%% file: sec/2_related.tex
\section{Related Work}
\subsection{Adaptive Articulated Object Manipulation}
Constructing simulation datasets for articulated object manipulation has been extensively studied~\cite{xiang2020sapien,Mo_2019_CVPR, Chang2015ShapeNetAI, geng2023gapartnet,liu2022akb,urakami2019doorgym,li2024unidoormanip}. 
Most previous works are restricted to simple, non-adaptive manipulation mechanisms.
In contrast, AdaManip~\cite{wang2025adamanip} introduces five distinct adaptive mechanisms in simulation, covering a broader range of real-world articulated objects. Given its comprehensive adaptive capabilities, we adopt AdaManip as the primary testing environment for our framework.

Alongside simulation environments, extensive research has been conducted on articulated object manipulation algorithms, spanning affordance-based~\citep{Mo_2021_ICCV, wu2022vatmart,wu2023learningenv,wang2021adaafford,ning2023where2explore,ling2024articulated}, flow-based~\citep{eisner2022flowbot3d, zhang2023flowbot++, li2024flowbothd}, part-based~\citep{geng2023partmanip,geng2023sage}, RL-based~\citep{geng2022end} and diffusion-based~\cite{Chi2023DiffusionPV,Ze20243DDP,wang2025adamanip} methods. For adaptive manipulation, AdaManip~\cite{wang2025adamanip} collects adaptive demonstration to train an in-category adaptive diffusion policy. SAGE~\cite{geng2023sage} leverages VLM and GAPartNet~\cite{geng2023gapartnet} to map semantic parts to actionable parts for executing manipulation policies with interactive adjustment.

\subsection{Foundation Models for Robotics}
With the remarkable generalization ability of foundation language and vision models, numerous studies have explored their integration into robotics.

One line of research focuses on mimicking the success of LLM by collecting massive datasets~\cite{fang2023rh20t, walke2023bridgedata, brohan2022rt, o2024open, gong2023arnold} to train large vision-language-action (VLA) models\cite{driess2023palm, team2024octo, kim2024openvla, li2024cogact, liu2024rdt, liu2024robomamba, black2024pi_0}. Another approach leverages existing foundation models for reward function specification in reinforcement learning\cite{ma2022vip, ma2023eureka, mahmoudieh2022zero, cui2022can, alakuijala2023learning, zeng2024learning}.

A different branch of research explores hierarchical policies, where high-level reasoning is performed by vision-language models (VLMs) that invoke low-level control functions to generate manipulation trajectories~\cite{ahn2022can, huang2022language, huang2022inner, singh2023progprompt}. Code as Policies\cite{liang2023code} applies LLMs to generate robot control code, while VoxPoser\cite{huang2023voxposer} produces robot trajectories by generating value maps based on foundation models.

Recent studies SAGE\cite{geng2023sage} and CoPa\cite{huang2024copa} adopt object parts as an intermediate representation to constrain the policy, aligning with our approach. However, our method introduces a distinct part grounding pipeline and a part-affordance model to enable primitive skill execution, differentiating it from prior work.

%% file: sec/3_method.tex
\section{Method}

In this section, we introduce the three key components of AdaRPG, as illustrated in \cref{fig:fig1}.

\subsection{Part Dataset and Affordance Learning}
\input{fig-tab/part_num}
\input{fig-tab/fig2}

As discussed in \cref{intro} and illustrated in \cref{fig:fig1}, parts serve as a powerful intermediate representation for achieving significant generalization across object categories. Therefore, we aim to train a part affordance model that captures local geometric features for downstream execution.

Existing methods and datasets~\cite{wu2022vatmart,Mo_2021_ICCV} primarily focus on whole-object affordances. To address this limitation, we first construct and annotate a dataset with point-level affordance on parts. PartNet-Mobility~\cite{Mo_2019_CVPR,xiang2020sapien} is a comprehensive dataset of articulated objects with diverse geometries and categories. We select 11 categories of object from PartNet-Mobility and extract key functional parts, such as handles, buttons, and knobs. To ensure broad coverage, we carefully collect parts with diverse shapes and functionalities. The dataset statistics are summarized in \cref{tab:part_num}. Note that our dataset only contains the \textbf{part point clouds} from PartNet-Mobility, without the full object context. The statistics in \cref{tab:part_num} are provided solely to illustrate the source object categories from which these parts are extracted. 
Therefore, no complete shapes have been included during training.

Given a part point cloud, we use an automated algorithm to annotate affordances on its surface. Human annotation is only required to specify the part’s bounding box and affordance surface plane, based on the part’s function characteristics. For instance, for a bottle cap, the annotation focuses on the top surface while ignoring the sides. Once the relevant surface is identified, high-affordance points are automatically computed based on the part’s center and bounding box, completing the annotation. These high-affordance regions (highlighted in red) are centered around the part’s center, representing areas associated with higher manipulation success rates. The part center is determined through a two-stage process, involving centroid estimation from the point cloud and subsequent refinement using the bounding box center. As the affordance surface defines the orientation plane for the gripper, its points can be directly used as end-effector targets without additional transformation.

The resulting dataset with point-level affordance on parts is represented as $D_{pa}=\{(O_i,p_i,r_i)\}$, where $O_i$ is the part point cloud, $p_i$ is the selected point, $r_i$ is the actionability score, where $r_i=1$ (positive) or $r_i = 0$
(negative) denote the annotated interaction outcome. We visualize several representative parts and corresponding affordance annotations in \cref{fig:fig2}.

Following where2act~\cite{Mo_2021_ICCV} and VAT-MART~\cite{wu2022vatmart}, we employ Pointnet++~\cite{qi2017pointnet++} as our affordance model encoder backbone, extracting per-point dense feature over object parts, followed by a MLP and a Sigmoid function mapping the output to actionability score $V(O_i,p_i)\in[0,1]$. The training loss of affordance learning is the Binary Cross Entropy Loss:

\begin{equation}
  L_V = \text{BCELoss}_{D_{pa}}(r_i,V(O_i,p_i))
  \label{eq:loss}
\end{equation}

The affordance model 
$V$ is trained on our proposed dataset, which includes diverse geometric and functional parts. As a result, it can predict the affordance of extracted parts from novel objects during online inference, provided they are correctly grounded and segmented. In the next section, we will elaborate on the process of grounding and segmenting these parts, ensuring accurate affordance prediction for previously unseen objects.

\subsection{Part Grounding \& Segmentation}
\input{fig-tab/method1}

Since part grounding and segmentation are essential prerequisites for affordance inference, ensuring their accuracy across all potential objects during online interaction is crucial. To enhance their performance, we incorporate foundation vision and language models, which are pretrained on extensive, diverse datasets. This large-scale pretraining equips them with strong generalization capabilities, enabling them to recognize and interpret a wide range of object parts and features more effectively.

As illustrated in \cref{fig:method1}, the camera captures RGB images and sends them to GPT-4o~\cite{hurst2024gpt}. We prompt GPT-4o to identify the scene and object, then describe the functional part in no more than three sentences while ignoring the robot arm and carefully distinguishing different object parts. For example, in \cref{fig:method1}, the pressure cooker has two handles: the lower handle is fixed to the cooker body and is non-actionable, whereas the upper handle is movable and thus relevant for interaction.

GPT-4o generates a detailed part description, which is then used as an input prompt for the GroundingDINO~\cite{liu2024grounding,ren2024grounded} model to generate a bounding box for the corresponding part. Notably, our extensive experimental trials confirm that providing a detailed description as a prompt significantly improves GroundingDINO’s performance compared to using single-word prompts.

The bounding box generated by GroundingDINO serves as the input prompt for Segment Anything~\cite{ravi2024sam,kirillov2023segment}, which refines the segmentation by precisely delineating the target part from the surrounding object structure. This process ensures that only the relevant actionable component is isolated while filtering out extraneous details. The obtained segmentation mask is then back-projected onto the depth image, enabling the reconstruction of an accurate 3D part representation in the form of a segmented point cloud. This step is crucial, as it allows for the preservation of spatial and geometric information, which is essential for downstream affordance prediction.

The segmented part point cloud is then processed by the pre-trained affordance model, which generates a dense actionability score map, guiding downstream primitive skills such as grasping and rotating.

As depicted in \cref{fig:method1}, all foundation models (GPT, GroundingDINO, and SAM) operate in a frozen, training-free manner, requiring no additional fine-tuning. Instead, they rely on extensive pretraining on large-scale datasets to provide high-quality part identification, localization, and segmentation. The only trainable component in our framework is the affordance model, which is specifically optimized on our proposed dataset. This strategic combination of powerful pre-trained models with a task-specific affordance predictor enables our approach to achieve both broad generalization and task-adaptive precision.

\subsection{Primitive Skill as Function}

Once we obtain the affordance score of the part point cloud, we identify actionable interaction points by selecting those with scores above a predefined threshold $\epsilon$. The end-effector’s translation pose is then computed as the average position of these selected points. To determine the gripper orientation, we estimate the normal direction of the interaction point on the point cloud surface, ensuring an optimal grasp pose. This process forms the primitive grasp function, which serves as the initial step for further manipulation.

Beyond grasping, we implement six atomic manipulation functions centered on the robot’s end-effector, defined along two principal axes of the end-effector coordinate frame: the z-axis (aligned with the gripper’s pointing direction) and the y-axis (parallel to the gripper’s two fingers and perpendicular to the z-axis). For z-axis, two translation actions (push/pull) move the end-effector along its pointing direction, while two self rotational actions allow the gripper to rotate around its own axis. For y-axis, the two translation actions refer to moving the part around the object axis. Each atomic action function returns a success flag upon completion, serving as feedback from the environment. For example, the \textit{pull\_part} function returns success when the end-effector’s movement reaches a predefined step size.

To ensure smooth and stable execution, all movements are governed by impedance control, which dynamically adjusts force and compliance to enhance interaction stability. By defining atomic actions in the end-effector’s local frame, our approach enables flexible and adaptive manipulation of objects in diverse orientations. The translation-based actions move the end-effector by a fixed step in the specified direction, while the rotational actions execute controlled rotations. To maintain precision, we continuously adjust the end-effector by computing and correcting angle deviations between its actual and expected positions in real-time. This correction mechanism, combined with impedance control, improves robustness across diverse object geometries and interaction scenarios, making the primitive skills more robust functions for high-level programs.

\subsection{High-Level Code Generation}

With the primitive skill functions, we leverage foundation model reasoning to generate high-level Python control code. Using natural language input, GPT-4o generates structured code that dynamically orchestrates primitive skills such as grasping, pulling, and rotating to adaptively manipulate articulated objects. All code is produced from a single general prompt and follows a unified generation process, ensuring scalability across object categories.

The generated code example follows an adaptive control loop, as shown in \cref{fig:method2}. The process begins with a grasp action, followed by repeated rotation attempts to adjust the part's position. A probability-based condition, which increases with the number of rotations, determines whether a pull attempt is made. If successful, the part is considered unlocked, and the process transitions to a secondary loop that continues executing the \textit{pull\_part} function until the task is complete. This approach ensures that the robot persistently adapts to unexpected constraints, increasing robustness in manipulation.

By leveraging foundation models to generate structured control logic, our method enhances generalization across different articulated objects, eliminating the need for manually scripted task-specific policies. The resulting code can be seamlessly integrated into real-world robotic systems, demonstrating the effectiveness of our framework in mechanism reasoning and high-level control strategies.

\input{fig-tab/method_reason}

%% file: fig-tab/part_num.tex
\begin{table*}[t]
\centering
\setlength{\tabcolsep}{2mm}
\begin{tabular}{cccccccccccc}
\toprule
\textbf{Categories} & \textbf{Window} & \textbf{Bottle} & \textbf{CM} & \textbf{Door} & \textbf{Safe} & \textbf{Faucet} & \textbf{Pot} & \textbf{Microwave}& \textbf{Pen} & \textbf{Switch}& \textbf{Toaster}\\
\toprule
\textbf{Part Count} & 4  & 14  & 4 & 20 & 20 & 14 & 15 & 7 &11 & 7 & 8\\
\bottomrule
\end{tabular}
\caption{\textbf{Statistics} of our Part Affordance Dataset. Note that our dataset only contains the \textbf{part point clouds} derived from PartNet-Mobility, without the full object context. The part counts indicate how many parts are extracted from each category.  \textbf{CM} stands for coffee machine.
}
\label{tab:part_num}
\end{table*}

% \begin{table*}
% \centering
% \begin{tabular}{|c|c|c|}
% \hline
% 列1 & 列2 & 列3 \\
% \hline
% 数据1 & 数据2 & 数据3 \\
% 数据4 & 数据5 & 数据6 \\
% \hline
% \end{tabular}
% \caption{这是一个横跨双栏的表格}
% \label{tab:my_table}
% \end{table*}

%% file: fig-tab/fig2.tex
\begin{figure}
  \centering
  \includegraphics[width=0.45\textwidth]{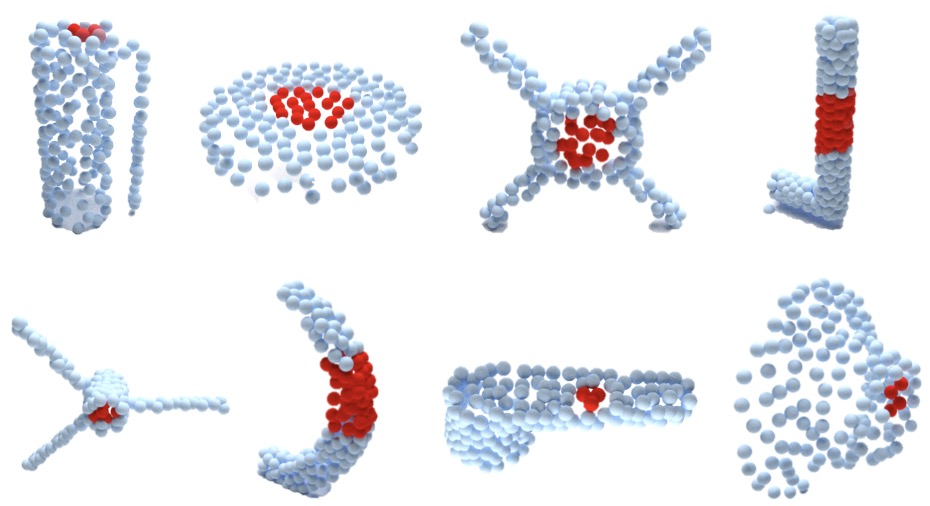}
  \caption{Visualization of several representative parts and affordance annotations in our Part Affordance Dataset.
  }
  \label{fig:fig2}
  \vspace{-0.3cm}
\end{figure}

%% file: fig-tab/method1.tex
\begin{figure}
  % \centering
  \includegraphics[width=0.5\textwidth]{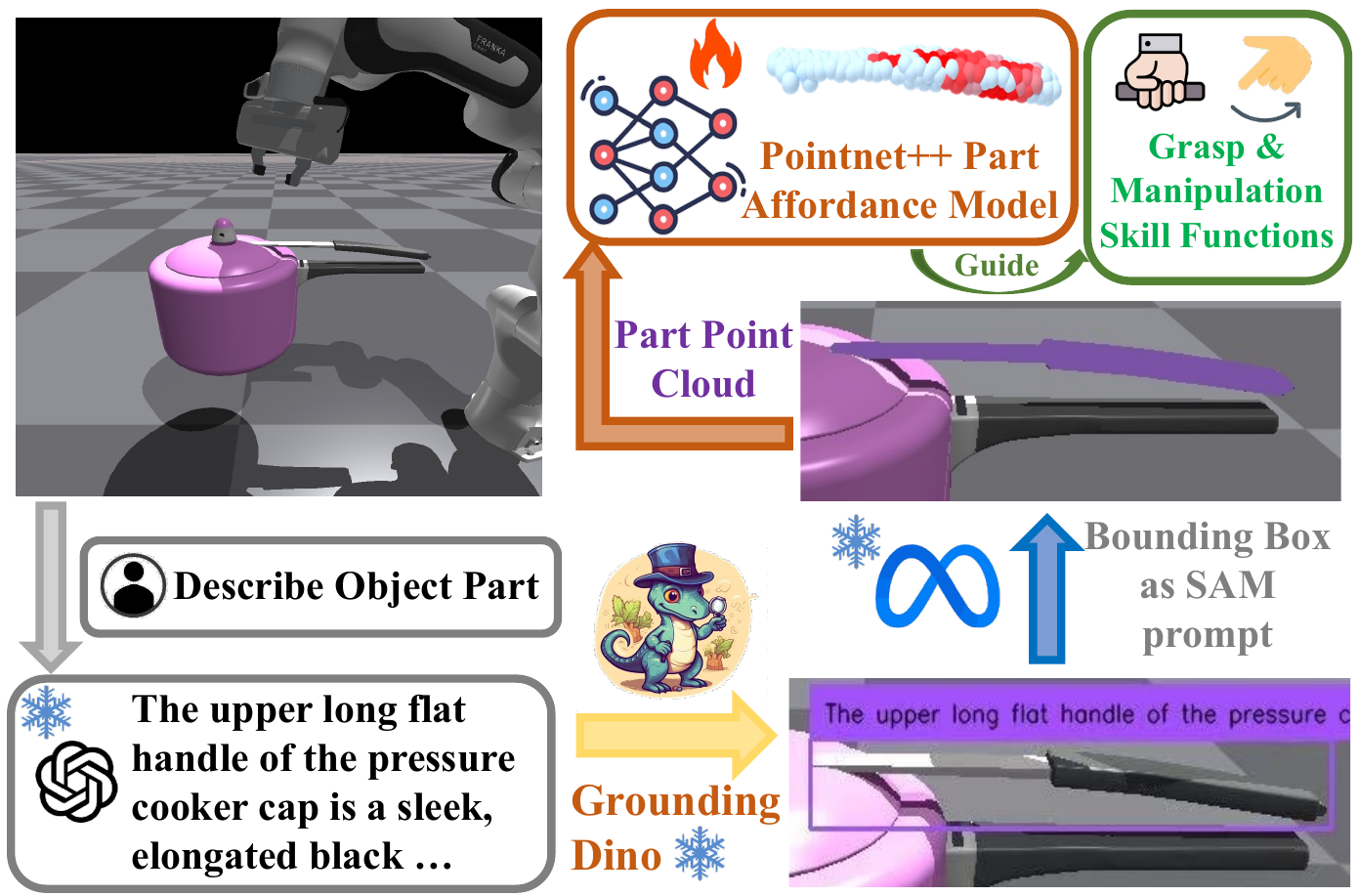}
  \caption{Foundation Models Guided Part Grounding and Segmentation. The RGB image is first processed by GPT-4o to generate a detailed part description, which is then used as an input prompt for GroundingDINO. GroundingDINO generates a bounding box for the identified part, which serves as the input prompt for Segment Anything to perform precise segmentation. The segmented part's point cloud is subsequently fed into the pre-trained part affordance model to infer an affordance score, guiding downstream primitive skills for manipulation.}
  \label{fig:method1}
  % \vspace{-0.3cm}
\end{figure}

%% file: fig-tab/method_reason.tex
\begin{figure}
  % \centering
  \includegraphics[width=0.5\textwidth]{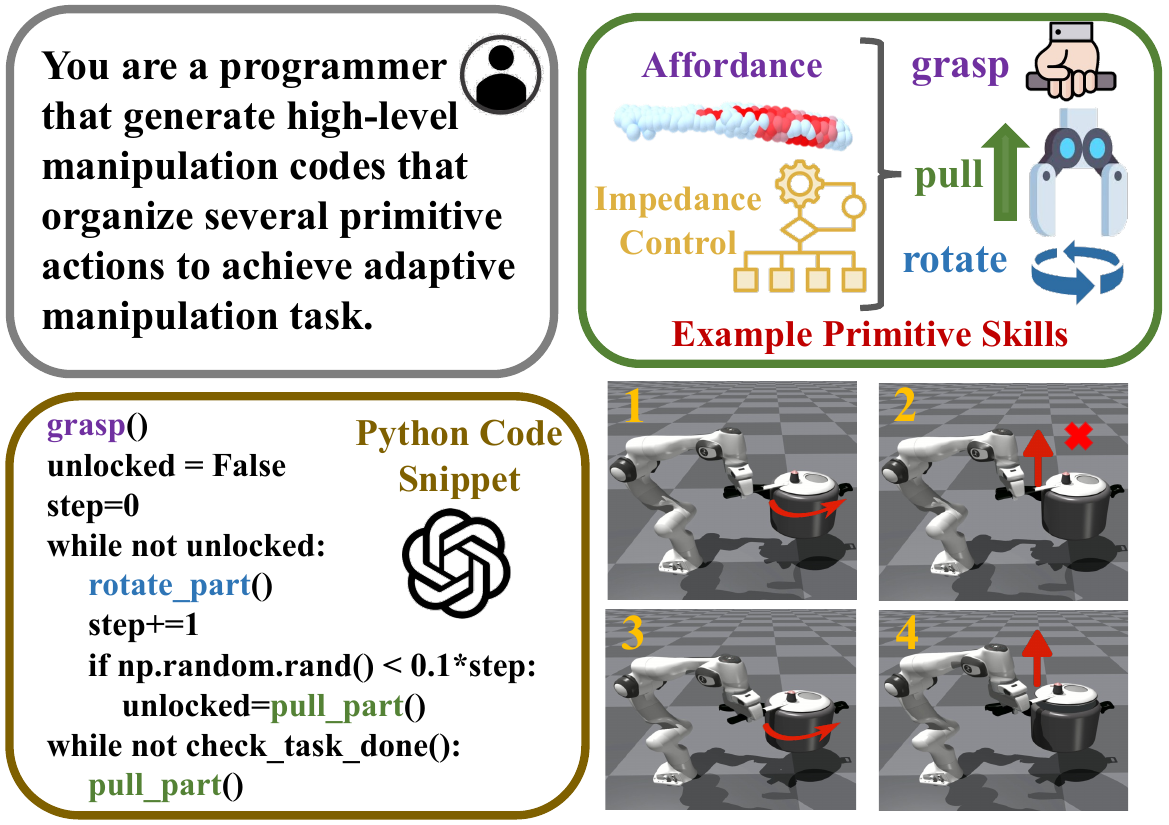}
  \caption{Foundation Model Reasoning and High-Level Python Control Code Generation. The lower-left image displays a Python snippet example generated by GPT-4o, which orchestrates adaptive manipulation by calling the grasp, pull, and rotate part functions. The code forms a while loop, ensuring the robot continues rotating if the pulling action fails, enabling iterative refinement for successful manipulation.}
  \label{fig:method2}
  \vspace{-0.3cm}
\end{figure}

%% file: sec/4_experiments.tex
\section{Experiments}

\subsection{Simulation Settings}
We perform cross-category generalization experiments using the AdaManip simulation environment\cite{wang2025adamanip}, which is built on IsaacGym\cite{makoviychuk2021isaac}. Our evaluation spans seven distinct object categories: bottle, pen, pressure cooker, coffee machine, window, door, and lamp, covering a wide range of articulated mechanisms and functional variations.

A key aspect of our evaluation is that our affordance model is not trained on the AdaManip dataset, and our part affordance training set derived from PartNet-Mobility lacks object-level annotations. This ensures that our framework is tested on entirely \textbf{novel object categories}, making it an unbiased assessment of its generalization capability.

To quantitatively evaluate performance, we report IoU between predicted and ground-truth part segmentation masks, F1 score for part affordance prediction, and manipulation success rate across all object categories. These metrics assess both individual components of our framework and the overall policy’s ability to generalize to previously unseen articulated objects. Achieving a high success rate requires not only accurate grounding of part segmentation and affordance, but also the ability to generate structured control code that reliably sequences primitive skills to complete tasks in a stable and efficient manner. Together, these evaluations highlight the framework’s capacity for complex, adaptive interaction with diverse articulated objects.

\subsection{Baselines and Ablations}
\label{baseline}

To thoroughly evaluate the effectiveness of our proposed method in handling complex adaptive manipulation tasks, we conduct a comprehensive comparison with state-of-the-art approaches leveraging large language models (LLMs) and imitation learning (IL) relying on expert adaptive demonstrations to infer manipulation policies. Additionally, we include two ablated versions of our method to evaluate the contribution of individual components.

 While LLM-based methods excel in semantic reasoning and task decomposition, they often struggle with precise execution due to limited physical understanding and lack of real-world affordance grounding. In contrast, IL-based methods are proficient in replicating expert behaviors, but their performance is constrained by dataset diversity and difficulty in adapting to novel objects.

\begin{itemize}
    \item \textbf{VAT-MART}\citep{wu2022vatmart}, an affordance-based method that predicts open-loop manipulation actions. Here we use VAT-MART solely as a baseline for affordance prediction.
    \item \textbf{SAGE}\citep{geng2023sage}, maps semantic parts of articulated objects to Generalizable Actionable Parts (GAParts) using a vision-language model and a domain-specific model GAPartNet\cite{geng2023gapartnet} to generate executable manipulation policies from natural language instructions with interactive feedback for robustness. 
    \item \textbf{CoPa}\citep{huang2024copa}, a robotic manipulation framework that leverages foundation vision-language models to generate 6-DoF end-effector poses by integrating task-oriented grasping and task-aware motion planning. It selects grasping parts using a coarse-to-fine grounding mechanism and derives post-grasp poses by identifying spatial geometry constraints of task-relevant object parts. 
    \item \textbf{AdaManip}\citep{wang2025adamanip}, an imitation learning approach that trains a diffusion policy using rule-based adaptive demonstration data. Since the original AdaManip algorithm is designed for in-category generalization, we extend its training to include all seven object categories. However, the test objects remain entirely novel to the training set.
    \item \textbf{Ours w/o prompt}. Part segmentation masks are generated without GPT-4o descriptions, using a generic prompt for GroundingDINO instead.
    \item \textbf{Ours w/o affordance}. The affordance module is removed, and manipulation points are set to the center of the segmented part, effectively treating all regions as equally actionable (i.e., assigning uniform affordance scores).

\end{itemize}

\subsection{Simulation Results}

\subsubsection{Part Segmentation}

We begin by evaluating the robustness of the foundation models in part grounding and segmentation, which constitutes the first stage of our inference pipeline. We use Intersection over Union (IoU) as the evaluation metric, comparing the predicted part segmentation masks, generated by our pipeline (GPT-4o + GroundingDINO + SAM), against ground-truth masks. As shown in Table~\ref{tab:iou}, the IoU exceeds 80\% across all object categories, demonstrating reliable performance in part grounding and segmentation.
\input{fig-tab/mask_iou}

\subsubsection{Affordance Prediction}

We further evaluate the performance of our affordance prediction module, benchmarking against VAT-MART~\citep{wu2022vatmart}, an affordance-based method for articulated object manipulation. Both models are trained on affordance dataset derived from PartNet-Mobility. While VAT-MART learns to predict affordance scores over the entire object point cloud, our model takes as input only the part point cloud.

During inference, VAT-MART predicts affordance scores for the full object, which we then mask using the ground-truth part segmentation to isolate the part-level prediction. In our approach, we directly extract the ground-truth part point cloud and predict affordance scores at the part level. Consequently, all evaluations are conducted on part-level affordance predictions. This allows us to ablate the affordance prediction performance from the influence of upstream part segmentation.

To assess cross-category generalization, we evaluate both methods on seven object categories from the AdaManip dataset using the F1-score metric. As shown in Table~\ref{tab:affordance}, our model significantly outperforms VAT-MART. We attribute this to VAT-MART’s reliance on whole-object geometry, which introduces high geometric variability across categories. In contrast, our part-level formulation reduces geometric variance, enabling more robust and generalizable affordance prediction.

\input{fig-tab/affordance_result}

\subsubsection{Object Manipulation}

\input{fig-tab/baseline}

We compare AdaRPG with baselines and ablations on the final object manipulation success rates. We present the quantitative results in \cref{tab:task}, which demonstrate that AdaRPG outperforms all baselines and ablations across all object categories. Notably, while AdaManip is trained on a larger dataset covering all seven categories, SAGE, CoPa, and AdaRPG remain entirely unexposed to the test data, making AdaRPG’s superior generalization even more compelling.

Despite being VLM-powered zero-shot manipulation frameworks, both SAGE and CoPa fall behind AdaRPG due to key limitations in their part representation and execution strategies. SAGE relies on GAPartNet for part grounding, which processes the entire point cloud and predicts part poses for rule-based primitive actions. However, pose estimation lacks the precision of affordance-based representations, leading to unstable execution as errors in part type estimation propagate through the manipulation pipeline.

On the other hand, CoPa incorporates GraspNet~\cite{fang2020graspnet}, a foundation grasp model trained on large-scale grasping data. While GraspNet excels at generic object grasping, our tasks require functional grasping, where the robot must interact with specific parts to complete sequential manipulation tasks. In practice, most grasp candidates generated by GraspNet fail to align with the actionable parts of articulated objects, leading to failures early in execution.

In contrast, AdaRPG utilizes a part affordance model to generate a fine-grained actionability score, enabling precise impedance-control-based primitive skills. By integrating multiple foundation models for part grounding and reasoning, AdaRPG achieves significantly higher success rates, demonstrating superior generalization and stability across diverse articulated objects.

To further analyze our method, we perform an ablation study. As shown in \cref{tab:task}, removing prompt guidance for part segmentation results in failures due to inaccurate part grounding. The affordance module significantly improves contact point selection, leading to a 15\% higher average success rate compared to the w/o-affordance baseline, highlighting its strong generalization across parts.
\input{fig-tab/sim_result}

Additional qualitative results are visualized in \cref{fig:sim_fig}, visualizing AdaRPG’s performance across three selected object categories. The results demonstrate that foundation-guided part grounding and affordance-based generalization effectively identify actionable parts, ensuring stable and precise manipulation. This strong part segmentation and affordance reasoning provide a robust foundation for high-performance execution, enabling AdaRPG to generalize well to novel articulated objects.
\input{fig-tab/real_result}

\subsection{Real-World Experiments}
\input{fig-tab/real_quant_baseline}

To evaluate the real-world generalization of our framework, we conduct experiments on a diverse set of everyday articulated objects, including a pressure cooker, microwave, bottle, and lamp. For these experiments, we utilize a Franka Emika Panda robotic arm equipped with an Intel RealSense D415 depth camera.

Following our framework, we first extract the part mask from the RGB image using foundation models, ensuring accurate part segmentation. The segmented part point cloud is then processed to predict per-point affordance scores, identifying the most actionable regions for interaction. The point with the highest affordance score is selected as the optimal grasping location, after which the robot executes primitive manipulation skills such as pulling, rotating, or pushing to complete the task. This pipeline enables adaptive and precise real-world interaction, demonstrating the robustness of our affordance-based approach in handling diverse articulated objects.

The quantitative results of real-world experiments are presented in \cref{tab:real_quant}, showing that all three baselines underperform compared to AdaRPG. Notably, by comparing \cref{tab:real_quant} with \cref{tab:task}, we observe that the real-world performance of AdaRPG surpasses simulation results. This improvement can be attributed to the fact that all foundation models are trained on real-world data, whereas the simulation environment introduces a domain gap that slightly hinders performance. This finding suggests that incorporating more foundation models into a robotic system could further enhance its ability to handle real-world challenges, leveraging their strong generalization capabilities to bridge perception and control effectively.

We present qualitative results from real-world experiments in \cref{fig:real_fig}, demonstrating the effectiveness of our approach in part recognition and segmentation across all four object categories. The results confirm that our foundation-guided part grounding pipeline successfully identifies and isolates actionable components, enabling precise manipulation. Additionally, the affordance model trained on simulation data transfers effectively to real-world point clouds, reinforcing the robustness and adaptability of affordance-based representations. This successful transfer further validates our design choice, highlighting the advantages of part-level affordance modeling over alternative representation strategies in generalizing across diverse environments.

%% file: fig-tab/mask_iou.tex
\begin{table}[!ht]
\vspace{-5pt}
  \centering
  % \scriptsize
  \setlength{\tabcolsep}{1mm}
  \resizebox{0.48\textwidth}{!}{
  \begin{tabular}{ccccccccccc}
     \toprule
    \textbf{Categories}&
\textbf{Bottle}&  
\textbf{Pen}& 
\textbf{PC}&
\textbf{CM}& 
\textbf{Window}& 
\textbf{Door}& 
\textbf{Lamp}\\
 \midrule
\textbf{IoU} &
0.99  &  
0.97  & 
0.84  &  
0.87  & 
0.88   &
0.99    &
0.80  \\

    \bottomrule
  \end{tabular}}
  \vspace{-0.2cm}
  \caption{\textbf{IoU between predicted and GT part masks.} PC and CM stand for Pressure Cooker and Coffee Machine, respectively. The same abbreviations are used in the following tables.}
\label{tab:iou}
\vspace{-0.3cm}
\end{table}

%% file: fig-tab/affordance_result.tex
\begin{table}[!htb]
\vspace{-5pt}
  \centering
  \small
  \setlength{\tabcolsep}{1mm}
  \resizebox{0.46\textwidth}{!}{
  \begin{tabular}{cccccccccc}
     \toprule
   \textbf{Method} &
\textbf{Bottle}&  
\textbf{Pen}& 
\textbf{PC}&
\textbf{CM}& 
\textbf{Window}& 
\textbf{Door}& 
\textbf{Lamp}\\
 \midrule
VAT-MART\citep{wu2022vatmart} &
0.56  &  
0.18  & 
0.21  &  
0.29  & 
0.34  &
0.17  &
0.16  \\

\midrule
\textbf{Ours} &
\textbf{0.82}  & 
\textbf{0.66}  & 
\textbf{0.72}  & 
\textbf{0.83}  &
\textbf{0.91} & 
\textbf{0.61}  &
\textbf{0.64} \\
    \bottomrule
  \end{tabular}}
  \vspace{-0.1cm}
  \caption{\textbf{Part Affordance F1 score comparison.}}
\label{tab:affordance}
\vspace{-0.1cm}
\end{table}

%% file: fig-tab/baseline.tex
% \usepackage{multirow}
% \begin{table}[tb]
% \begin{center}
% {
% \begin{tabular}{c|ccccccccc}

\begin{table}[t]
  \centering
  % \scriptsize
  \setlength{\tabcolsep}{1mm}
  \resizebox{0.48\textwidth}{!}{
  \begin{tabular}{cccccccccc}
    \toprule
    \textbf{Task} & \multicolumn{7}{c}{\textbf{Adaptive Manipulation}}                   \\
    % \cmidrule(c){1-2}
    % Name     & Description     & Size ($\mu$m) \\
    % \midrule
    % Dendrite & Input terminal  & $\sim$100     \\
    % Axon     & Output terminal & $\sim$10      \\
     \toprule
   \textbf{Method} &
\textbf{Bottle}&  
\textbf{Pen}& 
\textbf{PC}&
\textbf{CM}& 
\textbf{Window}& 
\textbf{Door}& 
\textbf{Lamp}\\
 \midrule
SAGE\cite{geng2023sage}   &
0.21  &  
0.40  & 
0.00  &  
0.30  & 
0.38   &
0.39    &
0.40  \\
CoPa\cite{huang2024copa} & 
0.58    & 
0.47    & 
0.17    & 
0.40   & 
0.08   &
0.39    &
0.20    \\

AdaManip\cite{wang2025adamanip} & 
0.46 & 
0.53 & 
0.50  & 
0.60 & 
0.46 &
0.44 &
0.30 \\
\midrule
Ours w/o prompt& 
0.11    & 
0.00    & 
0.00    & 
0.00   & 
0.00   &
0.00    &
0.20    \\
Ours w/o affordance &
0.78  &  
0.58  & 
0.83  &  
0.60  & 
0.57   &
0.63    &
0.40  \\

\textbf{Ours} &
\textbf{0.84}  & 
\textbf{0.73}  & 
\textbf{1.00}  & 
\textbf{0.80}  &
\textbf{0.84} & 
\textbf{0.78}  &
\textbf{0.70} \\
    \bottomrule
  \end{tabular}}
  \caption{\textbf{Object Manipulation Success Rates} of baselines, ablations and our method in simulation. Our proposed AdaRPG outperforms baseline and ablation methods in all categories.}
\label{tab:task}
\vspace{-0.2cm}
\end{table}

%% file: fig-tab/sim_result.tex
\begin{figure}
  \centering
  \includegraphics[width=0.5\textwidth]{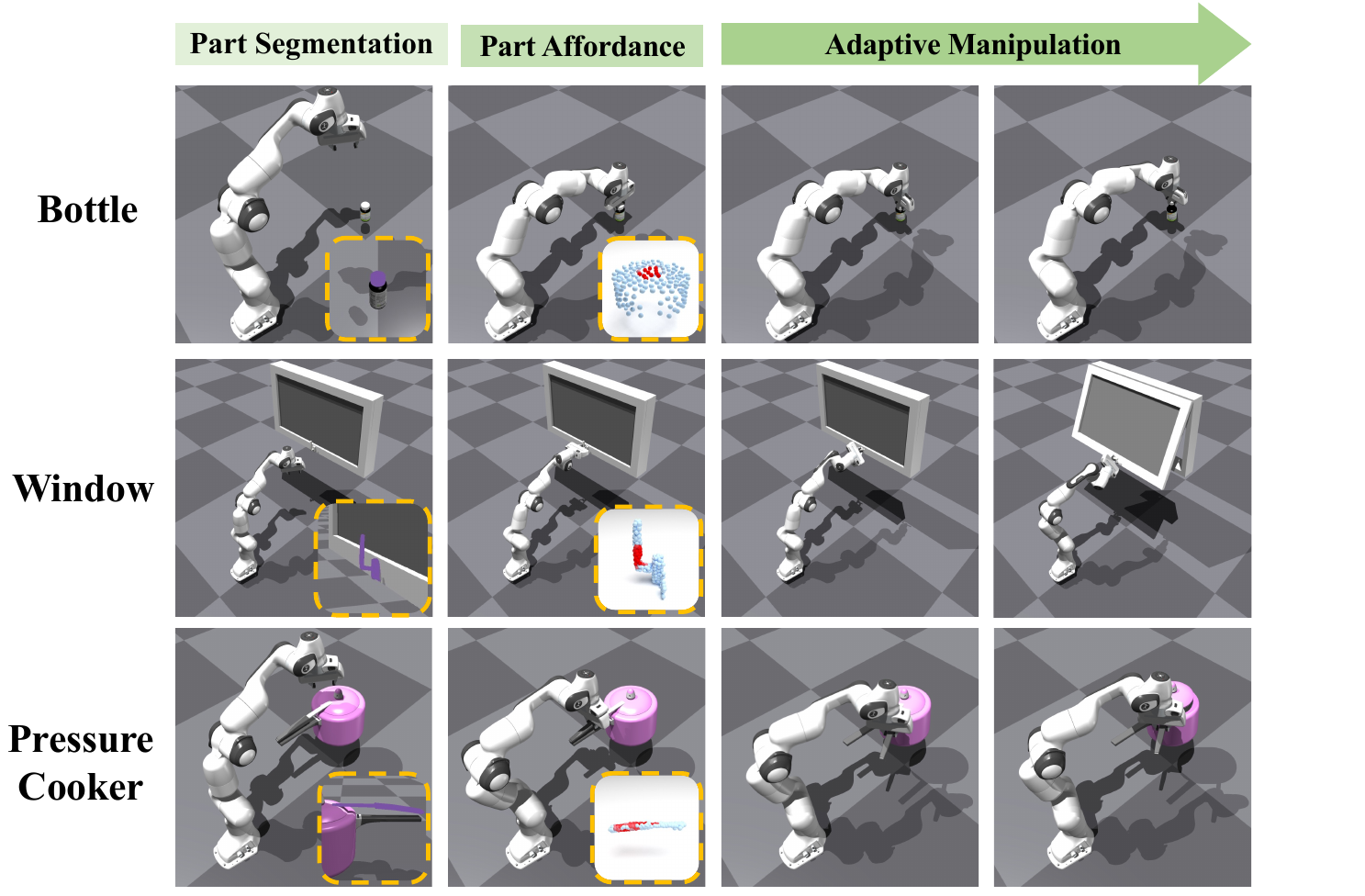}
  \caption{\textbf{Qualitative Simulation Result.} The intermediate results of part segmentation and affordance prediction demonstrate the effectiveness of AdaRPG.}
  \label{fig:sim_fig}
  \vspace{-0.1cm}
\end{figure}

%% file: fig-tab/real_result.tex
\begin{figure*}
  \centering
  \includegraphics[width=0.75\textwidth]{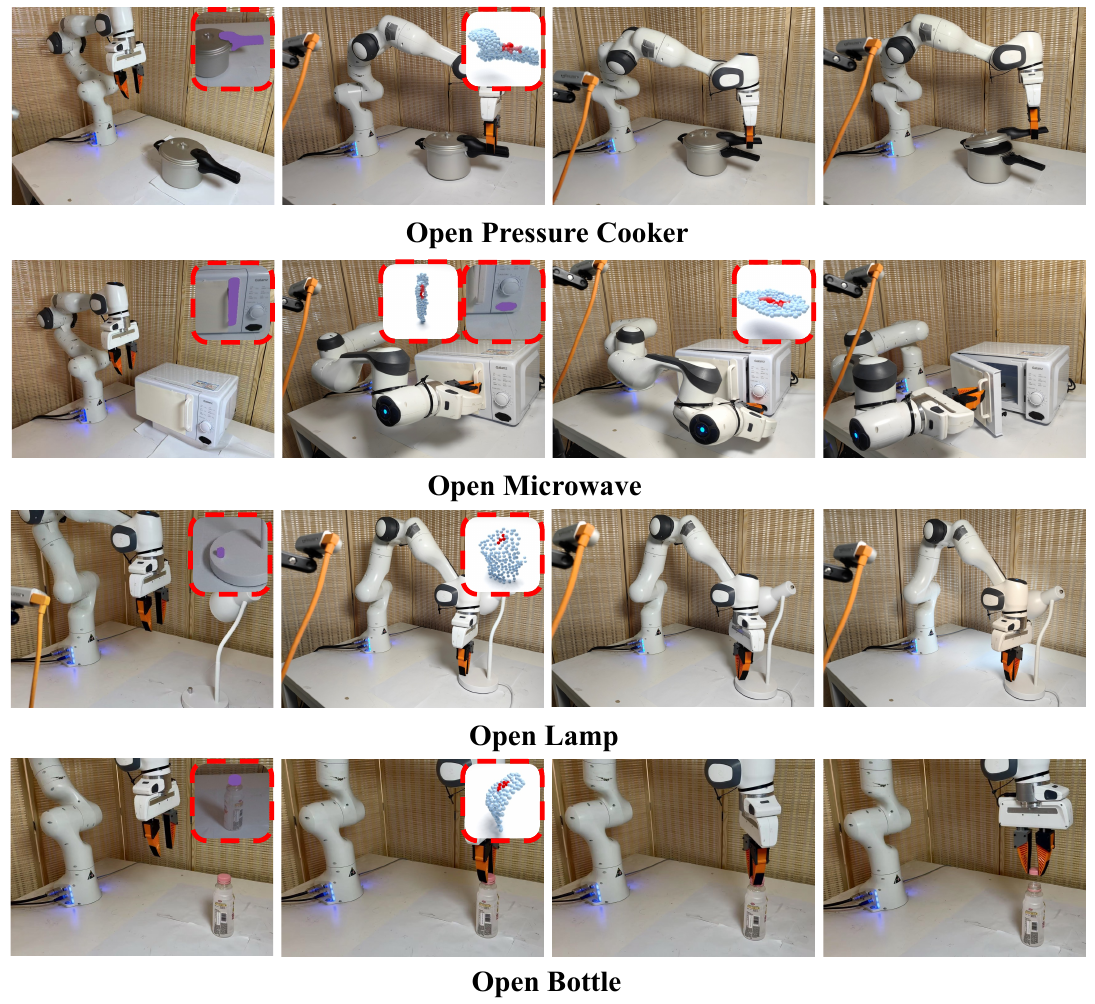}
  \vspace{-0.1cm}
  \caption{\textbf{Qualitative Real-World Results}.  We visualize the segmented handle, knob, and cap parts, along with their corresponding affordance score predictions. The high-level control codes leverage affordance-guided primitive actions to successfully complete the tasks.}
  \label{fig:real_fig}
  \vspace{-0.1cm}
\end{figure*}

%% file: fig-tab/real_quant_baseline.tex
\begin{table}[!htb]
\vspace{-5pt}
  \centering
  \small
  \setlength{\tabcolsep}{1mm}
  
  \begin{tabular}{cccccccccc}
     \toprule
   \textbf{Method} &
\textbf{Bottle}&  
\textbf{Pressure Cooker}&
\textbf{Microwave}& 
\textbf{Lamp}\\
 \midrule

SAGE\cite{geng2023sage} &
5/10  & 
6/10  & 
3/10  & 
6/10 \\

CoPa\cite{huang2024copa} &
4/10  &  
2/10 & 
3/10 &  
1/10 \\

AdaManip\cite{wang2025adamanip} &
8/10 &
5/10 &
7/10 &
5/10 \\

\midrule
\textbf{Ours} &
\textbf{9/10}  &
\textbf{10/10} & 
\textbf{9/10}  &
\textbf{8/10} \\

    \bottomrule
  \end{tabular}
  \vspace{-0.2cm}
  \caption{\textbf{Real-World Object Manipulation Success Rates.}}
\label{tab:real_quant}
\vspace{-0.2cm}
\end{table}

%% file: sec/5_conclusion.tex
\section{Conclusion}

In this work, we presented \textbf{AdaRPG}, a novel framework that leverages foundation models to enhance the generalization of adaptive articulated object manipulation by segmenting object parts and reasoning about their functional affordances. By shifting the focus from whole-object to part-level affordance inference, AdaRPG overcomes the challenges posed by geometric diversity and functional variability in real-world articulated objects. Our approach integrates a foundation model-powered pipeline with a structured affordance model, enabling adaptive primitive skill execution through high-level control code generation.

To support this framework, we introduced a part-level affordance annotation dataset, which provides a structured learning foundation for affordance inference, guiding precise manipulation strategies across diverse object categories. Experimental results in both simulation and real-world environments validate the robust generalization capability of AdaRPG, demonstrating its effectiveness in handling novel articulated objects beyond training data.

Moving forward, we aim to further enhance AdaRPG’s adaptability by integrating interactive affordance learning under dynamic environmental conditions. We hope that this work can pave the way for more generalizable and adaptable robotic manipulation strategies, bridging the gap between vision-language models and real-world robotic control.

%% file: sec/6_ack.tex
\section*{Acknowledgment}

This work was supported in part by the National Natural Science Foundation of China under Grants 62176025, 62301066, U21B2045, and 62206012.\\
\vspace{-0.4cm}